\documentclass[conference]{IEEEtran}
\IEEEoverridecommandlockouts

\usepackage{cite}
\usepackage{amsmath,amssymb,amsfonts}
\usepackage{algorithmic}
\usepackage{graphicx}
\usepackage{textcomp}
\usepackage{xcolor}
\def\BibTeX{{\rm B\kern-.05em{\sc i\kern-.025em b}\kern-.08em
    T\kern-.1667em\lower.7ex\hbox{E}\kern-.125emX}}

\usepackage{url}
\usepackage{xcolor}
\usepackage{amsmath}
\usepackage{xspace}
\usepackage{amssymb}
\usepackage{varwidth}
\usepackage{booktabs}
\usepackage{multirow}
\usepackage{cuted}
\usepackage{caption}
\usepackage[hidelinks]{hyperref} 
\usepackage[capitalize]{cleveref}

\makeatletter
\DeclareRobustCommand\onedot{\futurelet\@let@token\@onedot}
\def\@onedot{\ifx\@let@token.\else.\null\fi\xspace}

\def\ie{\emph{i.e}\onedot}

\def\etal{\emph{et al}\onedot}
\makeatother

\crefname{algocf}{algorithm}{algorithms}
\crefname{algocf}{Algorithm}{Algorithms}

\crefname{algocfline}{line}{lines}
\crefname{algocfline}{Line}{Lines}

\usepackage[ruled,vlined]{algorithm2e}
\SetKw{KwFrom}{from}
\SetKw{KwStep}{with step}
\SetKwInOut{Input}{input}
\SetKwInOut{Output}{output}
\SetKwComment{Comment}{}{} 

\def\DATASETNAME{\mbox{IRIS-v2}\xspace} 

\definecolor{darkgray}{rgb}{0.35,0.35,0.35} 

\begin{document}
\title{An Industrial Dataset for Scene Acquisitions and Functional Schematics Alignment
}


\author{
\IEEEauthorblockN{Flavien Armangeon$^{1,2}$, 
Thibaud Ehret$^3$, 
Enric Meinhardt-Llopis$^1$, \\
Rafael Grompone von Gioi$^1$, 
Guillaume Thibault$^2$, 
Marc Petit$^2$, 
Gabriele Facciolo$^{1,4}$}\\

\IEEEauthorblockA{
    \textit{$^1$ Universit\'e Paris-Saclay, ENS Paris-Saclay, CNRS, Centre Borelli, 91190, Gif-sur-Yvette}, France \\
    \textit{$^2$ EDF R\&D (Electricité de France)}, France  \quad \textit{$^3$  AMIAD, Pôle Recherche}, France \quad \textit{$^4$  Institut Universitaire de France} 
    }
}

\maketitle

\vspace*{-10pt}

\begin{strip}
\centering
\vspace*{-10pt}
\includegraphics[width=\linewidth]{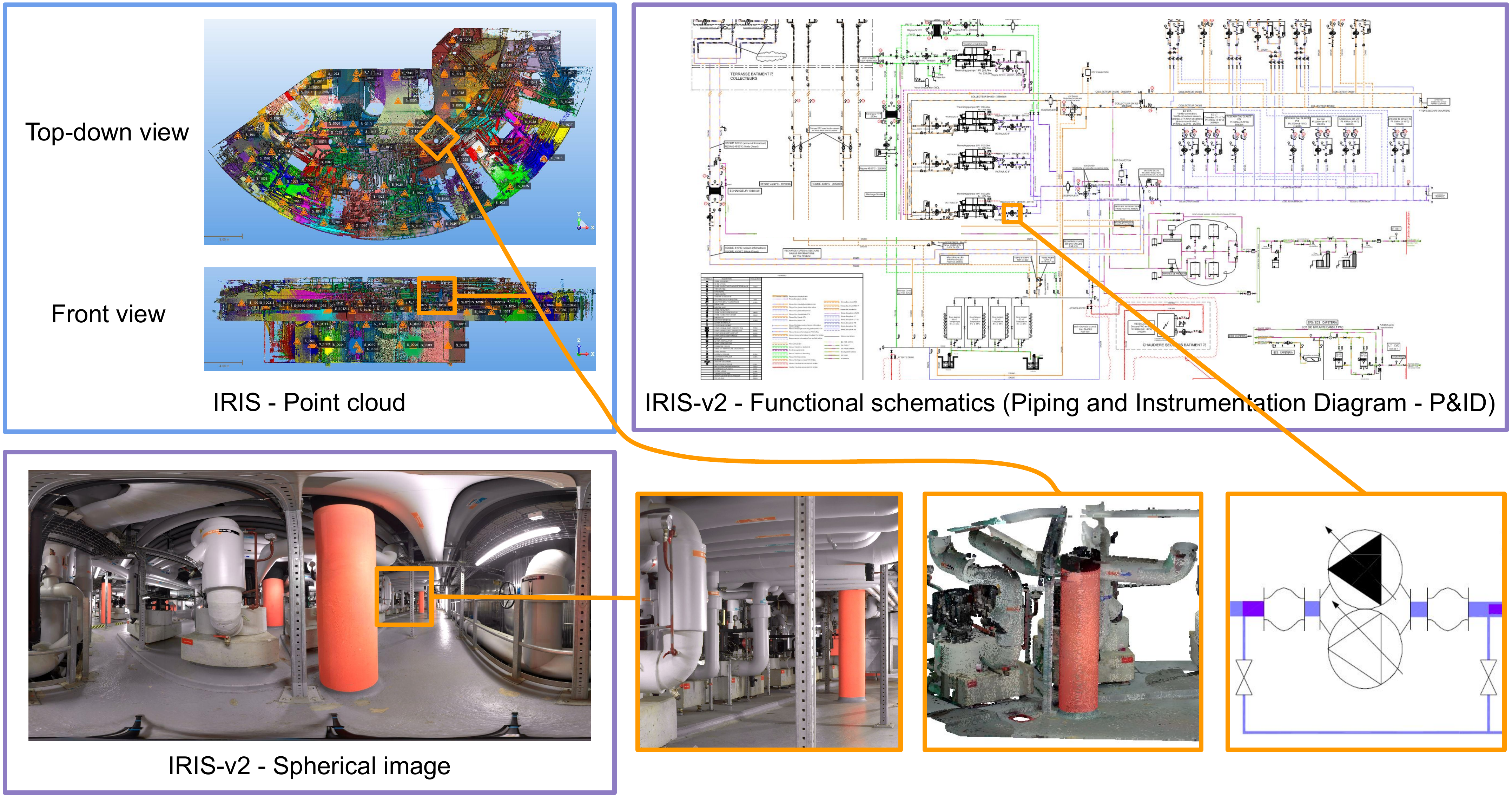}
\captionof{figure}{
This paper introduces \DATASETNAME (violet), a new dataset based on IRIS~\cite{armangeon2026iris} (blue) that provided a dense point cloud and a CAD model in a room exceeding $530 m^2$, and was used for point visibility estimation \cite{armangeon2025iris-vis}. In addition to IRIS data, \DATASETNAME introduces 300 spherical images, functional schematics with $\mathord{\sim}500$ pieces of equipment, pipe routing, 6000 annotated boxes and 47000 segmentation masks. These data are used for the problem of scene acquisitions and functional schematics alignment.
}
\label{fig:teaser}
\vspace*{-5pt}
\end{strip}

\begin{abstract}
Aligning functional schematics with 2D and 3D scene acquisitions is crucial for building digital twins, especially for old industrial facilities that lack native digital models. Current manual alignment using images and LiDAR data does not scale due to tediousness and complexity of industrial sites. Inconsistencies between schematics and reality, and the scarcity of public industrial datasets, make the problem both challenging and underexplored.
This paper introduces \DATASETNAME, a comprehensive dataset to support further research. It includes images, point clouds, 2D annotated boxes and segmentation masks, a CAD model, 3D pipe routing information, and the P\&ID (Piping and Instrumentation Diagram). The alignment is experimented on a practical case study, aiming at reducing the time required for this task by combining segmentation and graph matching.\footnote{\footnotesize Code and data: \url{https://centreborelli.github.io/scene-functional-alignment}}
\end{abstract}


\section{Introduction}
\label{sec:intro}
Pairing 3D scene with functional schematics is essential for creating digital twins. Both modalities provide complementary information: 3D data captures object geometry and pipe routing, while schematics describe control loops and functional relationships between pieces of equipment through the pipes. Digital twins enable key industrial applications such as predictive modeling, maintenance~\cite{tan2019machine}, and operator training in virtual reality~\cite{patle2019operator}.

Large facilities may contain tens of thousands of equipment pieces connected through hundreds of meters of piping. Manual alignment is time-consuming and requires a human expert who could be dedicated to other critical tasks. On the other hand, automatically aligning these data modalities is challenging due to the scale and complexity of industrial sites, inconsistencies between schematics and as-built scenes, missing distance information in schematics, complex object shapes, and occlusions. This problem is not well studied in the literature mainly due to the lack of industrial data. No end-to-end solution is given.

To bridge this gap, we publish the dataset \DATASETNAME and illustrates its usefulness through the alignment task using images, point clouds, and the P\&ID.
The proposed approach is divided in three main steps: 1. segmentation to locate objects in the 3D environment, 2. graph construction for both the scene and schematics, and 3. automatic graph alignment with human-assisted inconsistencies correction. Inconsistencies, or differences between the graphs, can come from segmentation errors, occlusions or objects missing in the P\&ID.
Thus the method is designed to be robust to structure perturbation and is validated on a real case study.

\cref{sec:relatedwork} reviews the state of the art for acquisition to schematics alignment, \cref{sec:dataset} presents the dataset, and \cref{sec:alignment_use_case} exhibits a use case solution to address the problem.

\section{Related Work}
\label{sec:relatedwork}

No complete end-to-end framework has been experimentally demonstrated for the scene geometry to functional alignment task, and no dataset is available providing both real acquisitions and schematics.
Previous works have described partial steps to pair schematics with a 3D scene~\cite{Sierla2020Towards,sierla2022Roadmap}. Key tools are identified, and combining them might lead to a solution. The process can be summarized as follows: digitalize the schematics to extract the functional relations, locate equipment and pipeline in the scene from acquisitions, pair both data modalities using a unified data structure, verify the pairing and correct the inputs in case of errors.

\vspace*{1mm}

\noindent\textbf{3D segmentation.}
Recent deep learning approaches for zero-shot segmentation often perform poorly in industrial environments due to the lack of data in the training sets. When possible, fine-tuning on a limited amount of data can mitigate this limitation~\cite{mae2025industrial}.
\cite{yang2023sam3d,xu2025sampro3d} use foundation models for 2D segmentation and 2d-to-3D masks projection. SAI3D~\cite{Yin2024sai3D} and OpenMask3d~\cite{takmaz2023openmask3d} align the 2D and 3D masks detected separately. For pipeline tracing, the literature is scarce and the practical exploitation of the results is often left to an operator. Some methods try to model explicitly the pipes via simple primitives or elbow and T/Y-junctions \cite{kawashima2012automatic,qiu2014pipe}. Other methods \cite{fu2013pipe,pang2015automatic} rely on pattern recognition and matching. More recently, the trend has shifted toward deep learning via convolutional neural networks \cite{cheng2020deeppipes,xie2023built}.

\vspace*{1mm}

\noindent\textbf{Common representation between scene acquisitions and functional schematics.}
\cite{son20153dreconstruction, rahul2019automatic} propose a tree structure in which each branch corresponds to a pipeline. This strategy cannot handle cycles in the pipe routing. In \cite{rantala2019applyinggraph}, Rantala~\etal compare different graph representations and conclude that representing equipment as nodes and pipes as edges gives the best matching performance to find similarities between two facilities. However, this is different from the scene acquisitions-schematics alignment problem, where we also need to match the pipes. Pipes must be represented as nodes, just like equipment.
Sierla~\etal~\cite{Sierla2020Integrating} proposed a common representation for the 3D geometry and the topology using a graph structure whose nodes are equipment pieces and pipe junctions, and edges are straight line pipes, but without experimenting the matching based on this representation.

\vspace*{1mm}

\noindent\textbf{Scene acquisitions to functional schematics alignment.}
Starting from a unified representation, Rantala~\etal~\cite{rantala2019applyinggraph} experimented with multiple graph matching algorithms based on the optimization of an objective function. However, their approach requires anchors in input, \ie a priori known input matchings.
Recently, Tang~\etal proposed SLOTAlign~\cite{tang2023robust}, a graph matching method that combines optimal transport and structure learning approaches, providing high robustness to structure perturbations between graphs. 

\section{The IRIS Dataset}
\label{sec:dataset}

\begin{figure}[t]
\centering
\includegraphics[trim=0 570 0 0 0, clip, width=1.0\linewidth]{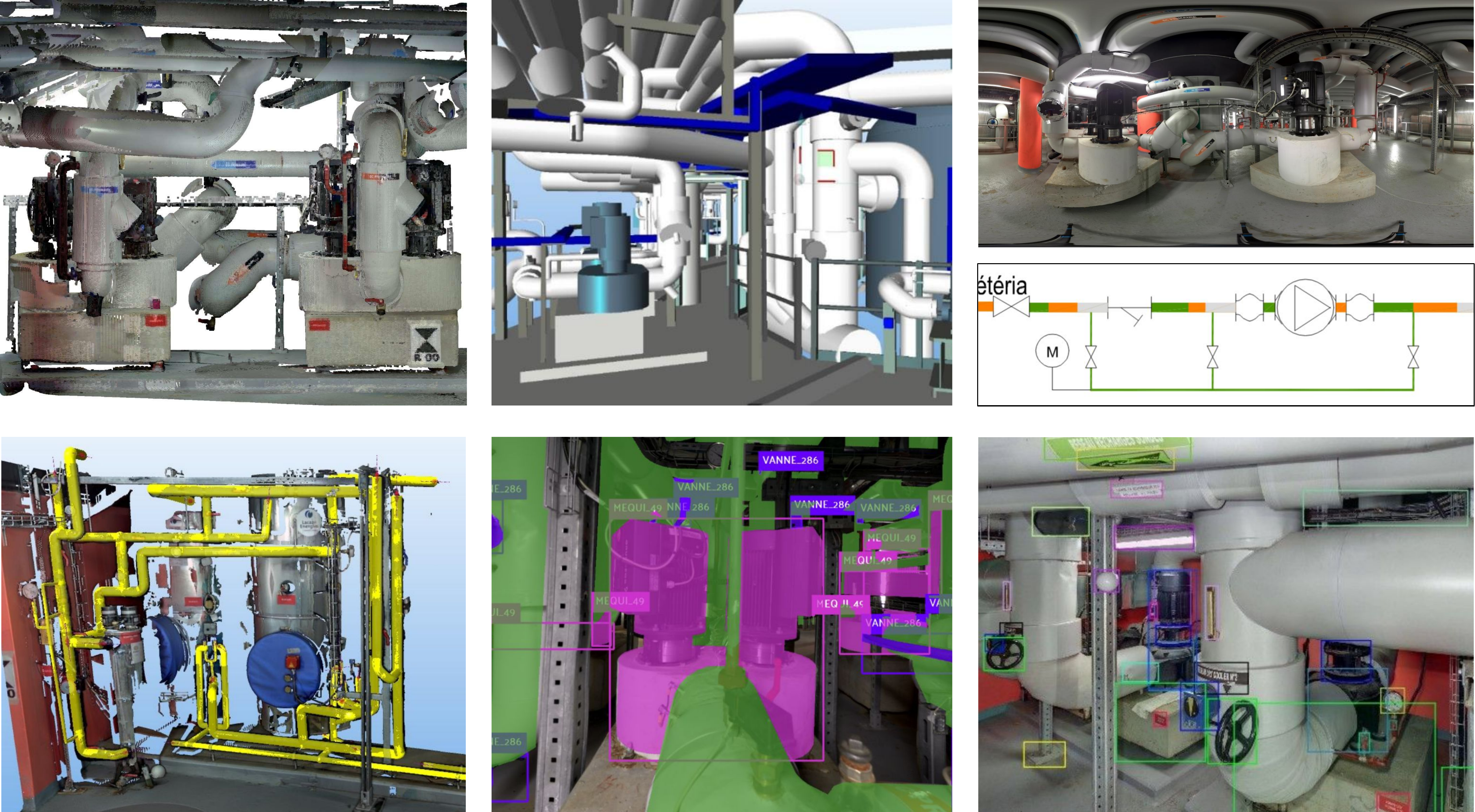} \\
(a) \hspace*{2.3cm} (b)  \hspace*{2.3cm} (c) \\
\includegraphics[trim=0 0 0 600, clip, width=1.0\linewidth]{images/resume_IRIS.pdf} \\
(d)  \hspace*{2.3cm} (e)   \hspace*{2.3cm} (f)
\caption{Data modalities of the \DATASETNAME dataset. (a) point cloud, (b) CAD model reconstructed close to the points, (c) 300 spherical images and the P\&ID, (d) pipelines routing, (e) annotated 2D masks, (f) annotated 2D boxes.}
\label{fig:resume_iris}
\vspace{-10pt}
\end{figure}

The \DATASETNAME dataset extends IRIS~\cite{armangeon2026iris}, a recently published dataset providing a dense point cloud and a CAD model in an industrial room of more than 530$m^2$. As shown in \cref{fig:teaser,fig:resume_iris}, this extension incorporates high resolution images, annotated bounding boxes, 2D segmentation masks for different categories, and the P\&ID, all paired with the point cloud. The  scene considered in IRIS is challenging for many computer vision tasks due to the complex shapes and the diversity of objects.

The dataset contains 300 spherical images of resolution 16384x8192 captured with an Xphase Pro X2 camera on a tripod at regular positions covering all the scene. Small objects are visible by far. These images are paired with a LiDAR point cloud merged from 67 acquisitions stations. The density is 150 points/cm² which is considered very high.

We publish two 3D models. First, the CAD model (\cref{fig:resume_iris}b) reconstructed close to the point cloud in a semi-automatic way with a tolerance of $3 \ \sigma = \pm 5 \ cm$. The objects are classified into 32 classes. Second, we make available the 3D pipe routing information extracted in a semi-automatic way from the point cloud using PipeRunner, a tool embedded in the RealWorks software published by Trimble. 
The software allows the user to successively pick points of the point cloud on a pipe and the reconstruction of elbows and bends between those points is completely automatic. The user thus progresses along the pipeline cloud, specifying when needed that there is a junction inducing a new branch.

We provide two annotated datasets. First, 2D human-annotated bounding boxes split into 171 classes of various objects and equipment classes (\cref{fig:resume_iris}f). It combines more than 6000 annotations.
Second, we publish 2D segmentation masks projected on the images from the 3D CAD model. It contains 47 000 annotations illustrated in \cref{fig:resume_iris}e.

The P\&ID, illustrated in \cref{fig:resume_iris}c, is provided in pdf format. Each individual equipment is represented as a symbol depending on its type and the pipes are lines between them.

\begin{figure}[t]
\centering
\includegraphics[trim=213 0 0 80, clip, width=0.96\linewidth]{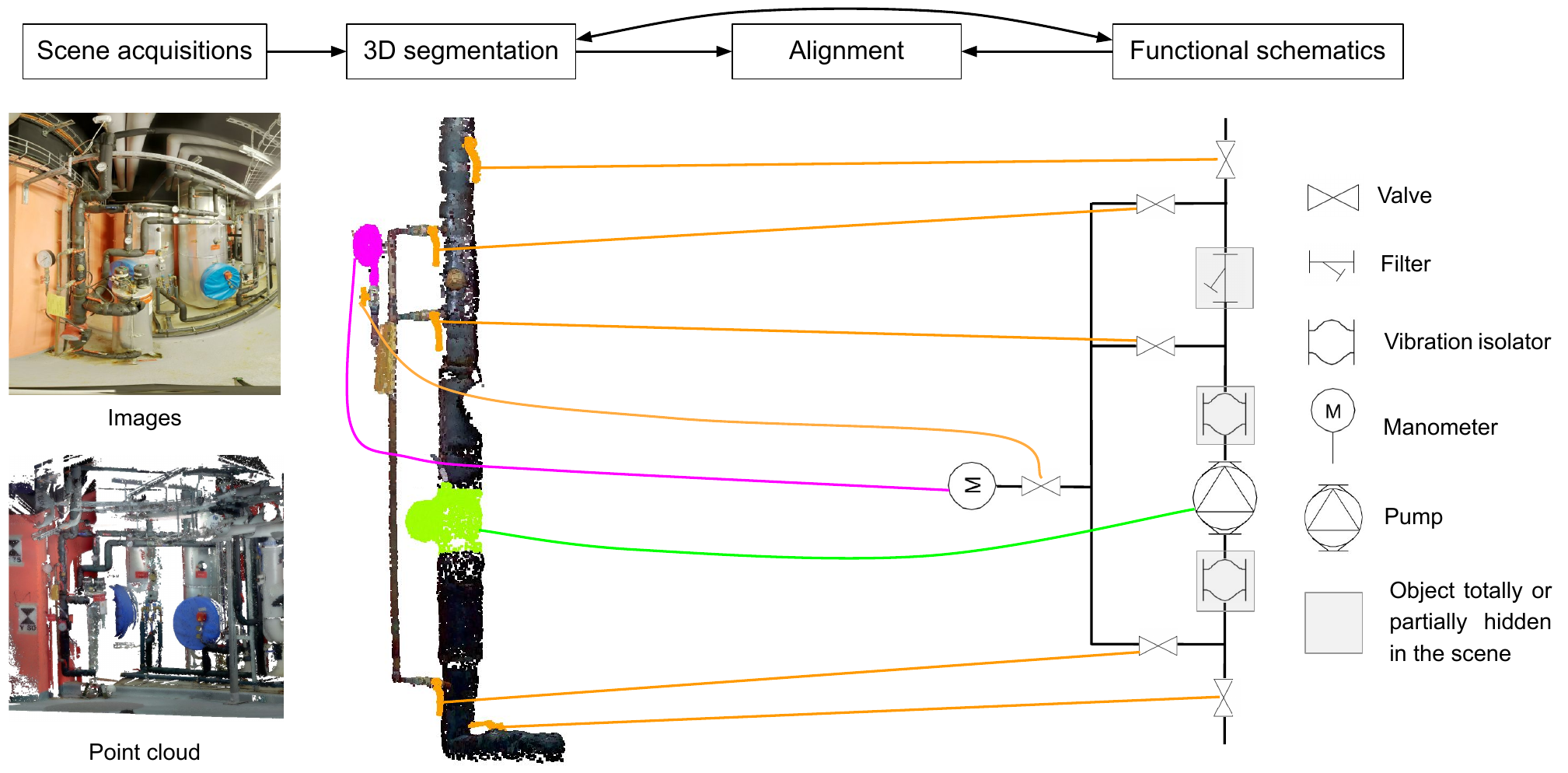}
\captionof{figure}{Example of alignment between scene acquisitions and functional schematics.  The problem consists in locating pieces of equipment from the functional schematics in the 3D scene. Since information on distances is not available in the schematics, the matching is done based on structure relationships between objects. The filter and vibration isolators cannot be segmented as they are hidden by heating isolation.
}
\label{fig:alignment_example}
\vspace*{-5mm}
\end{figure}

\cref{fig:alignment_example} illustrates the alignment problem which can be studied using this dataset. The objective is to pair regions of the scene with the P\&ID using 2D and 3D acquisitions. Due to occlusions, some objects are not visible in the images, nor in the point clouds.


\section{Scene-Functional Alignment Use Case}
\label{sec:alignment_use_case}

We show the usefulness of \DATASETNAME on the alignment task between acquisitions and schematics in the scene shown in \cref{fig:alignment_example}.
As illustrated in \cref{alg:cap}, the method consists in three main parts: segmentation, scene and functional graphs extraction, and nodes graph matching coupled with human resolution of the inconsistencies that are automatically detected.

\begin{algorithm}[htbp]
\caption{Proposed method}\label{alg:cap}
\KwIn{$S\!A$: scene acquisitions; $F\!S$: functional schematics}
\KwOut{$OC$: 3D object clouds; $M$: mapping}
\SetKwInput{KwVar}{Variables}
\KwVar{$S$: Scene graph; $F$: functional graph; $I\!N\!C$: inconsistencies}
$OC \gets \mathrm{3D\_segmentation}(S\!A)$\;
$S \gets \mathrm{Construct\_scene\_graph}(OC)$\;
$F \gets \mathrm{Construct\_functional\_graph}(F\!S)$\;
$M \gets \mathrm{Graph\_matching}(S, F)$ \tcp*[l]{S: source; F: target}
$I\!N\!C \gets \mathrm{Get\_inconsistencies}(M, S, F)$\;
\While{$I\!N\!C \neq \emptyset$}{
    $S, F \gets \mathrm{Human\_resolution}(I\!N\!C, M, S, F)$\;
    $M \gets \mathrm{Graph\_matching}(S, F)$\;
    $I\!N\!C \gets \mathrm{Get\_inconsistencies}(M, S, F)$\;
}

\end{algorithm}


\subsection{3D Segmentation}
\label{sec:method_segmentation}

First, we need to find each relevant object in the 3D scene. In order to have accurate information on the proximity between equipment and pipelines, fine grained 3D segmentation is required, especially for concave shapes such as pipe tees.

For equipment, similarly to SAM3D~\cite{yang2023sam3d}, we start by segmenting each image using 2D foundation models before projecting the segmentation masks onto the point cloud. Then, the 3D masks representing the same object are fused based on a minimum common points. This approach is more effective than directly segmenting the point cloud as current 3D foundation models do not perform well on industrial data, on the other hand it is much easier to obtain labeled images of industrial equipment to be used for fine-tuning.
We segment the images by first detecting the objects from an input text prompt using Grounding DINO~\cite{liu2023grounding} and then, by segmenting the main object in each box using SAM~\cite{kirillov2023segment}.
To improve the detection performance on industrial data, we finetuned Grounding DINO using our given dataset. Lastly, when projecting the 2D masks in 3D, some points belonging to objects behind the object of interest appear. Thus, we use the hidden point removal operator~\cite{katz2007direct} to remove the  hidden points before projecting. This technique is particularly useful in complex indoor environments such as industrial rooms \cite{armangeon2025iris-vis}.

Pipes are complex to segment. Unlike equipment, the dimensions and shapes of pipelines are not known in advance. The existence of junctions makes the automatization challenging in practice.
This is why we use PipeRunner, a semi-automatic piping lines reconstruction tool presented in \cref{sec:dataset} that demonstrated efficiency in performance and time ($> 200$ m/h). It includes the links between each pipe element, and features such as the type of pipe, the position, diameter, and extremities.

\begin{figure}[t]
\centering
\includegraphics[width=0.99\linewidth]{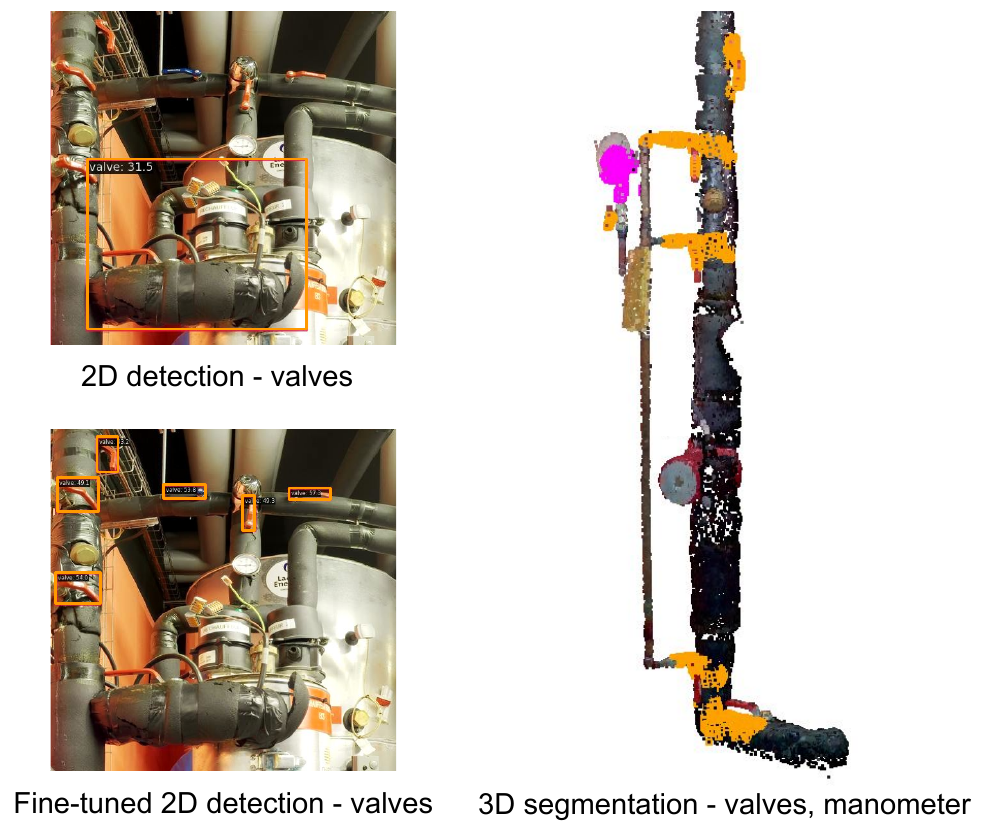}
\vspace*{-5mm}
\caption{2D detection and 3D Segmentation results using our fine-tuned Grounding DINO model~\cite{liu2023grounding} and SAM~\cite{kirillov2023segment} for 2D segmentation and standard 2D-to-3D masks projection and fusion for 3D segmentation.}
\label{fig:segmentation_results}
\vspace*{-5mm}
\end{figure}

Results are shown in \cref{fig:segmentation_results}. To improve the 2D detection performance on valves, we fine-tuned Grounding-DINO~\cite{liu2023grounding} on butterfly valves using our annotated boxes (\cref{sec:dataset}). In this example, all the valves are found except one. This valve is detected in other images so it is well segmented on the point cloud. The 2D boxes are then feed to SAM~\cite{kirillov2023segment} to obtain the 2D segmentation masks then projected to the point cloud. All pieces of equipment are accurately segmented in 3D except the pump which is hard to find, even for a human. This problem could be mitigated via fine-tuning if data is available. Thus, we segment the pump manually in this example.
The segmentation of pipelines, achieved using PipeRunner, is accurate. At the moment, fully automatic pipe segmentation cannot separate every single pipe element, such as cylinder or T-junction, needed to pair the P\&ID.

\subsection{Scene and Functional graphs construction}
\label{sec:geometrical_graph_construction}
\label{sec:topological_graph_construction}

To be aligned, the scene acquisitions and the functional schematics must be processed to extract the relevant information into a common representation. We propose a graph representation where each equipment and pipeline is a node and where edges represent objects that touch each other in the scene.
Pipelines are represented as nodes because they can be connected to more than two pieces of equipment and they need to be matched. They are cut at the intersections such as T or Y-junctions to keep the connection information that is shared for both data modalities.
An attribute is associated to each node depending on the type of the object.
To be sure that the representation is consistent at the boundaries between the scene and the functional topology, the pipeline nodes whose degree is lower than two (\ie open on one side) are removed.

\begin{figure}[t]
\centering
\vspace*{-1mm}
\includegraphics[trim=0 0 0 5, clip,width=\linewidth]{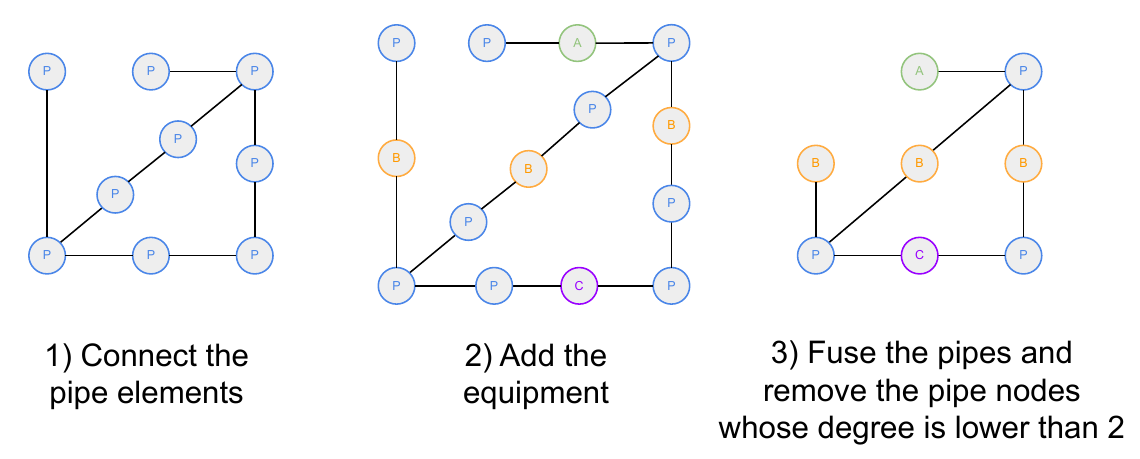}
\vspace*{-6mm}
\caption{Scene graph construction steps. P nodes are pipe elements and A, B and C nodes denote equipment types.}
\label{fig:geometrical_graph_construction}
\vspace*{-4mm}
\end{figure}

The scene graph $\mathcal{S}$ is extracted from the segmented objects in three steps, illustrated in \cref{fig:geometrical_graph_construction}.
First, the pipe elements are linked based on a threshold on their distances.
Each type of pipe is processed with suitable rules. A cylinder is connected to the two closer pipe elements, and a T or Y-junction is linked to the three closer elements. Secondly, for each piece of equipment, the distance is computed to the closest pipe element and if this distance is lower than the threshold, they are connected.
Then, the pipe nodes of degree two (ex: cylinders, elbow) that are connected to at least another pipe are removed, thus cutting the pipelines at junctions. Finally, the pipes at the end of a chain are removed.

From the schematics, the digitalized P\&ID is a graph where each equipment and pipe junction is a node. 
We obtain a graph that follows the representation illustrated in step 2 in \cref{fig:geometrical_graph_construction}. Thus, the last step is similar to the scene graph construction.

For the scene shown in \cref{fig:alignment_example}, we extract the scene graph $\mathcal{S}$ illustrated in \cref{fig:graph_matching_quali} with a threshold of 4cm on the distance between two pipe elements to be linked. If the hidden objects (filter and vibration isolators) are considered not important, thus removed from the schematics, the functional graph $\mathcal{F}$ is identical to $\mathcal{S}$ (\cref{fig:graph_matching_quali}-Source). This shows that our graph convention is shared between the scene and schematics. On the other side, if the filter needs to be paired with the scene, it is kept in $\mathcal{F}$ as shown in \cref{fig:graph_matching_quali}.

\subsection{Robust attributed graph matching}
\label{sec:method_matching}

The scene and functional graphs can be different in case of segmentation
errors, occlusions or objects that are not represented in the schematics. As the errors from the P\&ID are rare and due to human mistakes, $\mathcal{F}$ is more reliable than $\mathcal{S}$. Thus, we consider $\mathcal{S}$ as source graph and $\mathcal{F}$ as target graph.

We use SLOTAlign~\cite{tang2023robust} a graph matching method based on optimal transport that is able to use the node attributes and that has shown robustness capabilities to structure perturbation.

\begin{figure}[t]
\centering
\begin{tabular}{c}
\includegraphics[trim=0 0 0 0, clip, width=0.95\linewidth]{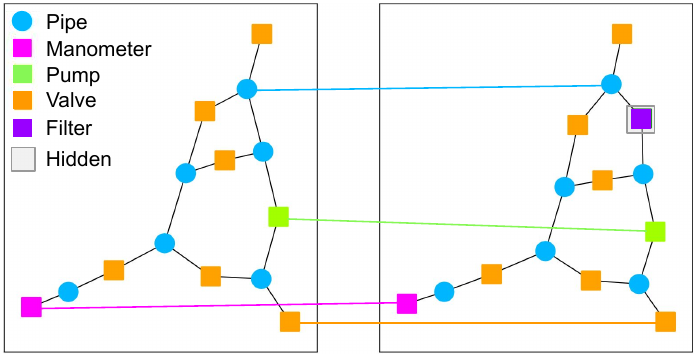}\\
\multicolumn{1}{l}{\hspace{0.3cm} Source graph (scene) \hspace{0.95cm} Target graph (schematics)}
\end{tabular}
\vspace*{-1mm}
\caption{Graph matching results of SLOTAlign~\cite{tang2023robust} on the scene illustrated in \cref{fig:alignment_example}. The filter in the schematics is hidden in the scene but kept in the schematics. The matching is perfect.}
\label{fig:graph_matching_quali}
\end{figure}

In \cref{fig:graph_matching_quali}, when the filter is not removed from the P\&ID, it is important to remark that  the alignment is still perfect. Thus, as the pipes connected to this filter are well paired with the P\&ID, we are able to locate approximately this object even though it is not visible in the scene. Here we see the importance of being robust to perturbations.

\subsection{Human resolution of inconsistencies}
\label{sec:human_resolution}

\begin{figure}[t]
\centering
\includegraphics[trim=0 0 0 0, clip, width=1.0\linewidth]{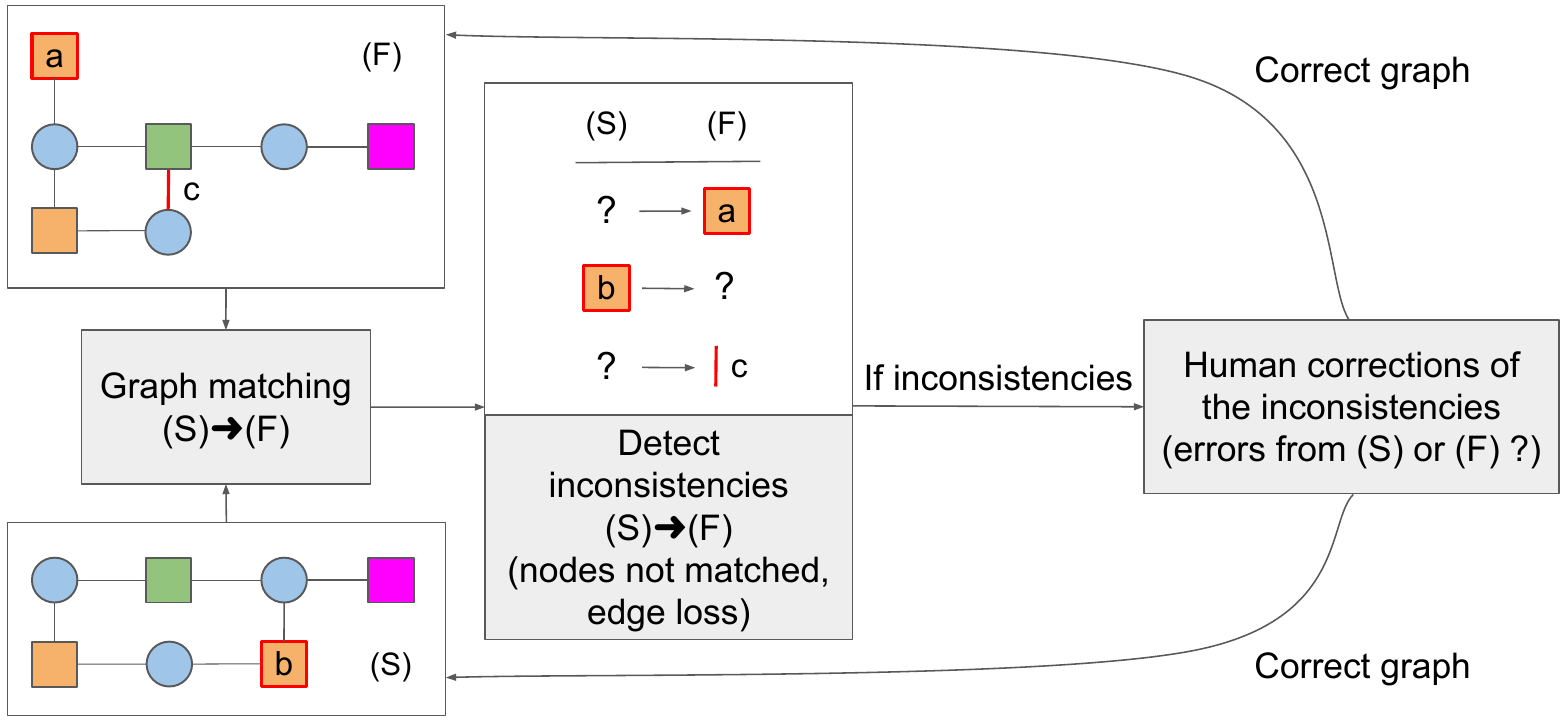}
\vspace*{-5mm}
\caption{Robust alignment: $\mathcal{S}$ is paired with $\mathcal{F}$ and the inconsistencies are detected and given to a human to correct the graphs. The process is repeated until no inconsistency is found.
}
\label{fig:method}
\vspace*{-4mm}
\end{figure}

As illustrated in \cref{fig:method}, the input graphs can contain inconsistencies, especially when modifications in the scene have not been reflected in the diagram. After graph matching, they are automatically detected and fed to a human for correction before repeating the matching step. The loop ends when no inconsistency is detected. These inconsistencies are of three types: nodes from $\mathcal{S}$ that are paired with the same node from $\mathcal{F}$, a node from $\mathcal{F}$ that has no preimage in $\mathcal{S}$ and an edge that not preserved through the mapping function.

\section{Conclusion}

We publish a new dataset in an industrial room and illustrate its usefulness via the scene acquisition to functional schematics alignment problem. The experiments on a practical case study show that the method can align both data modalities.
In the future, further experiments could be done on larger scenes included in IRIS or other domain if data become available. Also, one can imagine automatically tracing the pipes or correct the graphs inconsistencies.

\bibliographystyle{splncs04}
\bibliography{main.bib}


\end{document}